\documentclass[twoside]{article}
\usepackage{PRIMEarxiv}
\usepackage{graphicx}

\usepackage[utf8]{inputenc} 
\usepackage[T1]{fontenc}    
\usepackage{hyperref}       
\usepackage{url}            
\usepackage{booktabs}       
\usepackage{amsfonts}       
\usepackage{nicefrac}       
\usepackage{xcolor}         

\usepackage[numbers]{natbib}

\usepackage{amsopn}
\usepackage{epsfig}
\usepackage{ifthen}
\usepackage{algorithm}
\usepackage{algorithmic}
\usepackage{xcolor}

\usepackage{longtable}
\usepackage{amsmath}
\usepackage{amsfonts}
\usepackage{booktabs} 
\usepackage{xparse}
\usepackage{bigdelim}
\usepackage{caption}
\usepackage{subcaption}

\usepackage{cleveref}

\usepackage{etoolbox}
\newcommand{\ubold}{\fontseries{b}\selectfont}
\robustify\ubold

\renewrobustcmd{\bfseries}{\fontseries{b}\selectfont}
\renewrobustcmd{\boldmath}{}
\newrobustcmd{\B}{\bfseries}
\usepackage{siunitx} 
\newcommand{\vect}[1]{\boldsymbol{#1}} 
\newcommand{\mat}[1]{\boldsymbol{#1}} 
\newcommand{\tvect}[1]{\tilde{\boldsymbol{#1}}}
\newcommand{\tmat}[1]{\tilde{\boldsymbol{#1}}}
\newcommand{\tscal}[1]{\tilde{#1}}
\newcommand{\hvect}[1]{\hat{\boldsymbol{#1}}}
\newcommand{\hmat}[1]{\hat{\boldsymbol{#1}}}
\newcommand{\hscal}[1]{\hat{#1}}
\newcommand{\bvect}[1]{\bar{\boldsymbol{#1}}}
\newcommand{\bmat}[1]{\bar{\boldsymbol{#1}}}
\newcommand{\bscal}[1]{\bar{#1}}

\newcommand{\dummystring}{QWERTYU}
\newcommand{\vci}[3][\dummystr]{\ifthenelse{\equal{#1}{\dummystring}}{\vect{#2}_{#3}}{\vect{#2}_{#3}^{(#1)}}}
\newcommand{\mx}[3][\dummystr]{\ifthenelse{\equal{#1}{\dummystring}}{\mat{#2}_{#3}}{\mat{#2}_{#3}^{(#1)}}}
\newcommand{\tvci}[3][\dummystr]{\ifthenelse{\equal{#1}{\dummystring}}{\tvect{#2}_{#3}}{\tvect{#2}_{#3}^{(#1)}}}
\newcommand{\tmx}[3][\dummystr]{\ifthenelse{\equal{#1}{\dummystring}}{\tmat{#2}_{#3}}{\tmat{#2}_{#3}^{(#1)}}}
\newcommand{\tscl}[3][\dummystr]{\ifthenelse{\equal{#1}{\dummystring}}{\tscal{#2}_{#3}}{\tscal{#2}_{#3}^{(#1)}}}
\newcommand{\hvci}[3][\dummystr]{\ifthenelse{\equal{#1}{\dummystring}}{\hvect{#2}_{#3}}{\hvect{#2}_{#3}^{(#1)}}}
\newcommand{\hmx}[3][\dummystr]{\ifthenelse{\equal{#1}{\dummystring}}{\hmat{#2}_{#3}}{\hmat{#2}_{#3}^{(#1)}}}
\newcommand{\hscl}[3][\dummystr]{\ifthenelse{\equal{#1}{\dummystring}}{\hscal{#2}_{#3}}{\hscal{#2}_{#3}^{(#1)}}}
\newcommand{\bvci}[3][\dummystr]{\ifthenelse{\equal{#1}{\dummystring}}{\bvect{#2}_{#3}}{\bvect{#2}_{#3}^{(#1)}}}
\newcommand{\bmx}[3][\dummystr]{\ifthenelse{\equal{#1}{\dummystring}}{\bmat{#2}_{#3}}{\bmat{#2}_{#3}^{(#1)}}}
\newcommand{\bscl}[3][\dummystr]{\ifthenelse{\equal{#1}{\dummystring}}{\bscal{#2}_{#3}}{\bscal{#2}_{#3}^{(#1)}}}

\newcommand{\abs}[1]{\left|#1\right|}

\renewcommand{\eqref}[1]{Eq.~\ref{eq:#1}}

\newcommand{\qt}[1]{{p}_{#1}}
\newcommand{\alphat}[1]{{\alpha}_{#1}}

\newcommand{\xit}[1]{\mathbf{x}^{(i)}_{#1}}

\newcommand{\qit}[1]{{p}^{(i)}_{#1}}
\newcommand{\zit}[1]{z^{(i)}_{#1}}
\newcommand{\zthat}[1]{\hat{z}_{#1}}

\newcommand{\Zthat}[1]{\hat{Z}_{#1}}
\newcommand{\Zithat}[1]{\hat{Z}^{(i)}_{#1}}

\newcommand{\RPS}{\textrm{RPS}}
\newcommand{\coverage}{\mathrm{coverage}}
\newcommand{\indicator}[1]{\mathcal{I}_{[#1]}}

\newcommand{\model}[1]{\texttt{#1}}
\newcommand{\ETS}{\model{AutoETS}}
\newcommand{\ARIMA}{\model{AutoARIMA}}
\newcommand{\tbats}{\model{TBATS}}
\newcommand{\Thetaf}{\model{Theta}}
\newcommand{\Croston}{\model{Croston}}

\newcommand{\DeepAR}{\model{DeepAR}}
\newcommand{\SQF}{\model{SQF}}

\newcommand{\NPTS}{\model{NPTS}}
\newcommand{\SeasonalNPTS}{\model{Seas.NPTS}}
\newcommand{\Climatological}{\model{NPTS(uni.)}}
\newcommand{\SeasonalClimatological}{\model{Seas.NPTS(uni.)}}
\newcommand{\SeasonalNaive}{\model{Seas.Naive}}
\newcommand{\STLAR}{\model{STL-AR}}
\newcommand{\DeepNPTS}{\model{DeepNPTS}}
\newcommand{\GluonTS}{\model{GluonTS}}
\newcommand{\NA}{\model{NA}}

\newcommand{\dataset}[1]{\texttt{#1}}
\newcommand{\electricity}{\dataset{Electricity}}
\newcommand{\exchange}{\dataset{Exchange Rate}}
\newcommand{\traffic}{\dataset{Traffic}}
\newcommand{\taxi}{\dataset{Taxi}}
\newcommand{\wiki}{\dataset{Wiki}}
\newcommand{\solar}{\dataset{Solar Energy}}

\newcommand{\predlength}{{Pred. Len.}}
\newcommand{\domain}{{Domain}}
\newcommand{\freq}{{Freq.}}
\newcommand{\size}{{Size}}
\newcommand{\history}{{(Median) TS Len.}}
\newcommand{\numrolls}{{No. Windows}}
\newcommand{\numtestpoints}{{No. Test Points}}

\title{Deep Non-Parametric Time Series Forecaster}

\newcommand*\samethanks[1][\value{footnote}]{\footnotemark[#1]}

\author{
	Syama Sundar Rangapuram \\
	Amazon\\
	\texttt{rangapur@amazon.com} \\
	\And
	Jan Gasthaus\thanks{Work done while at Amazon.} \\
	\And
	Lorenzo Stella \\
	Amazon\\
	\texttt{stellalo@amazon.com} \\
	\AND
	Valentin Flunkert\\
	Amazon\\
	\texttt{flunkert@amazon.com} \\
	\And
	David Salinas \\ 
	Amazon\\
	dsalina@amazon.com \\ 
	\And
	Yuyang (Bernie) Wang\\
	Amazon\\
	\texttt{yuyawang@amazon.com} \\
	\And
	Tim Januschowski\samethanks[1] \\  
	Zalando\\
	\texttt{tim.januschowski@zalando.de} \\
}

\date{}

\begin{document}
	
\maketitle

\begin{abstract}
	This paper presents non-parametric baseline models for time series forecasting.
	Unlike classical forecasting models, the proposed approach does not assume any parametric form for the predictive distribution and instead generates predictions by sampling from the empirical distribution according to a \textit{tunable} strategy.
	By virtue of this, the model is always able to produce reasonable forecasts (i.e., predictions within the observed data range) without fail unlike classical models that suffer from numerical stability on some data distributions.
	Moreover, we develop a global version of the proposed method that automatically learns the sampling strategy by exploiting the information across multiple related time series. 
	The empirical evaluation shows that the proposed methods have reasonable and consistent performance across all datasets, proving them to be strong baselines to be considered in one's forecasting toolbox.
\end{abstract}


\section{Introduction}
Non-parametric models (or algorithms) are quite popular in machine learning for both supervised and unsupervised learning tasks especially in applied scenarios.\footnote{See for example \url{https://www.kaggle.com/kaggle-survey-2020}.} Examples of widely used, non-parametric models in supervised learning~\citep{Raschka} include the classical $k$-nearest neighbours algorithm, as well as more sophisticated tree-based methods like LightGBM and XGBoost~\citep{LightGBM17,Chen2016}.
In contrast to parametric models, which have a finite set of tunable parameters that do not grow with the sample size, non-parametric algorithms produce models that can become more and more complex with an increasing amount of data; for instance, decision surfaces learned by $k$-nearest neighbours or decision trees.
One of the main advantages of non-parametric models is that they work on any dataset, produce reasonable baseline results, and aid in developing more advanced models.

Despite the popularity of non-parametric methods in a general supervised setting, perhaps surprisingly, this popularity is not reflected similarly in the literature on time series forecasting.
Non-parametric methods based on quantile regression and bootstrapping have been explored in forecasting but are usually applied to intermittent time series data~\citep{petropoulos2021forecasting}.
Recent work~\citep{Wen2017multi, aistats} propose deep learning based extensions of quantile regression methods for forecasting.
However, much of the work in time series forecasting has traditionally been focussed on developing parametric models that typically assume Gaussianity of the data, e.g., ETS and ARIMA~\citep{Hyndman2008}.
Extensions of these classical models have been proposed to handle intermittent data~\citep{Croston1972}, count data~\citep{Snyder12}, non-negative data~\citep{Akram2008}, multi-variate distributions~\citep{multivariate,rasul2020multivariate} as well as more general non-Gaussian settings~\citep{seeger16,debezenac2020}. Unfortunately, most of these models cannot reliably work for all data distributions without running into numerical issues, which inhibits their usefulness in large-scale production settings. At the least, a robust, fail-safe model must be available to provide fall-back in case less robust methods exhibit erroneous behaviour~\citep{bose2017probabilistic}.

This paper therefore focuses on non-parametric methods for probabilistic time series forecasting that are practically robust. Our main contribution are novel non-parametric forecasting methods, which in more details are as follows.

\begin{itemize}
\item We propose a simple \textit{local} probabilistic forecasting method called Non-Parametric Time Series Forecaster (NPTS, for short) that relies on sampling one of the time indices from the recent \textit{context window} and use the value observed at that time index as the prediction for the next time step.\footnote{This is similar in spirit to $k$-nearest neighbours ($k$-NN) algorithm but unlike $k$-NN, our approach naturally generates a probabilistic output.} We further demonstrate how NPTS can handle seasonality by adapting its sampling strategy; trend can be handled via standard techniques such as differencing~\citep{hyndman2017forecasting}.
\item We propose a \emph{global} extension DeepNPTS, where the sampling strategy is automatically learned from multiple related time series. For this we rely on a simple feed-forward neural network that takes in past data of all time series along with (optional) time series co-variates and outputs the sampling probabilities for the time indices.
To train the model, we use a loss function based on the ranked probability score~\citep{Epstein:1969}, which is a discrete version of the continuous ranked probability score (CRPS)~\citep{matheson1976scoring}. Note that both RPS and CRPS are proper scoring rules for evaluating how likely the value observed is in fact generated from the given distribution~\citep{Epstein:1969, gneiting2007strictly}.
\end{itemize}
NPTS is robust by design and the robustness of DeepNPTS stems from the fact that once the model is trained, the sampling probabilities and then the predictions can be obtained by doing a forward pass on the feed-forward neural network, which does not suffer from any numerical problems.
In spite of being a deep-learning based model, DeepNPTS produces output that is explainable (Figure~\ref{fig:predictions}).
Moreover, it generates calibrated forecasts (Figure~\ref{fig:calib}) and for non-Gaussian settings like integer data or rate/interval data, NPTS in general and DeepNPTS in particular give much better results than the standard baselines which we show in extensive experiments by comparing against forecasting methods across the spectrum (local and global, parametric and non-parametric). Our experiments suggest that NPTS and DeepNPTS are indeed good and robust baseline forecasting methods. They  complement the limited amount of existing non-parametric forecasting methods. 

The article is structured as follows. We introduce the local model, NPTS, in Section~\ref{sec:npts} and then extend it to a global version called DeepNPTS in Section~\ref{sec:deep_npts}. After discussing related work in Section~\ref{sec:rel_work}, we provide both qualitative and quantitative experiments in Section~\ref{sec:experiments}. We conclude in Section~\ref{sec:conclusion}.

\section{Non-Parametric Time Series Forecaster}\label{sec:npts}
Here we introduce the non-parametric forecasting method for a single univariate time series. 
This method is \textit{local} in the sense that it will be applied to each time series independently, similar to the classical models like ETS and ARIMA.
In Section~\ref{sec:deep_npts}, we discuss a global version, which is more relevant to modern time series panels.

Let $z_{0:T-1} = (z_0, z_1, \ldots, z_{T-1})$ be a given univariate time series. The time series $z$ is univariate if each $z_i$ is a one-dimensional value. Note that we do not need to further specify the domain of the time series (e.g., whether $z \in \mathbb{R}^T$ or $z \in \mathbb{Z}^T$) as this is not important unlike for other methods. 
To generate prediction for the next time step $T$, NPTS randomly samples a time index $t \in \{0, \ldots, T-1\}$ from the past observed time range and use the value observed at that time index as the prediction.
That is,
\[   
    \hat{z}_T = z_t, \quad t \sim \qt{T}(\cdot), \quad  t \in \{0, \ldots, T-1\}, 
\]
where $\qt{T}(\cdot)$ is a categorical distribution with $T$ states.
To generate distribution forecast, we sample time indices from $p_T(\cdot)$ $K$ times and store the corresponding observations as Monte Carlo samples of the predictive distribution.

Note that this is quite different from naive forecasting methods that choose a fixed time index to generate prediction, e.g., $T-1$ or $T - \tau$, where $\tau$ usually represents a seasonal period.
The sampling of time indices, instead of choosing a fixed index from the past, immediately brings up two advantages:
\begin{itemize}
\item it makes the forecaster probabilistic and consequently allows one to generate prediction intervals,
\item it gives the flexibility by leaving open the choice of sampling distribution $\qt{T}$; one can define $\qt{T}$ based on prior knowledge (e.g., seasonality) or in more generally learn it from the data.
\end{itemize}

We now discuss some choices of sampling distribution $\qt{T}(\cdot)$ that give rise to various specializations of NPTS.
In Section~\ref{sec:deep_npts}, we discuss how to learn the sampling distribution from the data.

One natural idea for the sampling distribution is to weigh each time index of the past according to its ``distance'' from the time step where the forecast is needed. 
An obvious choice is to use the exponentially decaying weights as the recent past is more indicative of what is going to happen in the next step.
This results in what we call NPTS (without any further qualifications):
\[
        \qt{T}(t) \propto \exp(-\lambda \abs{T - t}),
\]
where $\lambda$ is a hyper-parameter that will be tuned based on the data.
We also refer to this variant as NPTS with exponential kernel.

\paragraph{Seasonal NPTS.}
One can generalize the notion of distance from simple time indices to time-based features $f(t) \in \mathbb{R}^D$, e.g., hour-of-day, day-of-week.
This results in what we call seasonal NPTS
\[
        \qt{T}(t) \propto \exp(-\lambda \abs{f(T) - f(t)}).
\]
The feature map $f$ can in principle be learned as well from the data. In this case, one keeps the exponential kernel intact but only learns the feature map.
The method presented in the next section directly learns $p_T$.

\paragraph{NPTS with Uniform Kernel (Climatological Forecaster).}

The special case, where one uses uniform weights for all the time steps in the context window, i.e., $\qt{T}(t) = 1/T$, leads to Climatological forecaster~\citep{gneiting2007strictly}.
One can similarly define a seasonal variant by placing uniform weights only on past seasons indicated by the feature map.
Again, this is different from the seasonal naive method~\citep{hyndman2017forecasting}, which uses the observation from the last season alone as the point prediction whereas the proposed method uniformly samples from several seasons to generate probabilistic prediction.

\paragraph{Extension to Multi-Step Forecast.} 
Note that so far we only talked about generating one step ahead forecast.
To generate forecasts for multiple time steps one simply absorbs the predictions for last time steps into the observed targets and then generates consequent predictions using the last $T$ targets. 
More precisely, let $\{\hat{z}_{T, k}\}_{k=1}^K$ be the prediction samples obtained for the time step $T$.
Then prediction samples for time steps $T + t,\ t > 0$ are generated auto-regressively using the values $(z_t, \ldots, z_{T-1},\hat{z}_{T, k}, \ldots, \hat{z}_{T+t-1, k}), k = 1, 2, \ldots, K$.


\section{Deep Non-Parametric Time Series Forecaster}\label{sec:deep_npts}
The main idea of DeepNPTS is to learn the sampling distribution from the data itself and continue to sample only from observed values. 
Let $N$ be a set of univariate time series $\{\zit{0:T_i - 1}\}_{i=1}^N$, where $\zit{1:T_i-1} = (\zit{1}, \zit{2}, \ldots, \zit{T_i-1})$ and $\zit{t}$ is a scalar quantity denoting the value of the $i$-th time series at time $t$ (or the time series of \emph{item} $i$).%
\footnote{We consider time series where the the time points are equally spaced, but the units or frequencies are arbitrary (e.g. hours, days, months). Further, the time series do not have to be aligned, i.e., the starting point $t=1$ can refer to a different absolute time point for different time series $i$.}
Further, let $\{\xit{0:T_i}\}_{i=1}^N$ be a set of associated, time-varying covariate vectors with $\xit{t} \in \mathbb{R}^D$. We can assume time-varying covariate vectors to be the only type of co-variates without loss of generality, as time-independent and item-specific features can be incorporated into $\xit{}$ by repeating the feature value over all time points.
Our goal is to learn the sampling distribution $\qit{T_i}(t), t = 0, \ldots, T_i - 1$ w.r.t. to the forecast start time $T_i$ for each time series.
The prediction for time step $T_i$ for the $i^{th}$ time series is then obtained by sampling from $\qit{T_i}$.

\subsection{Model}\label{sec:model}
Here we propose to use a feed-forward neural network to learn the sampling probabilities.
As inputs to this global model, we define a fixed length \textit{context window} of size $T$ spanning the last $T$ observations $z^{(i)}_{T_i-1 : T_i - T}$.
In the following, without loss of generality, we refer to this context window as the interval $[0, T-1]$ and the prediction time step as $T$; however, note that the actual time indices corresponding to this context window would be different for different time series.

For each time series $i$, given the observations from the context window, the network outputs the sampling probabilities to be used for prediction for time step $T$.
More precisely, 
\begin{equation}\label{eq:model}
        \qit{T}(t) = \Psi(\xit{0:T}, \zit{0:T-1}; \phi), \ t = 0, 1, \ldots, T-1.
\end{equation}
Here $\Psi$ is the neural network and $\phi$ is the set of neural network weights that are shared among all the time series.
The network outputs different sampling probabilities for each time series $i$, depending on its features.
However, these different sampling probabilities are parametrized by a single set of common parameters $\phi$, facilitating information sharing and global learning from multiple time series.

We assume from here onwards that the outputs of the network are normalized so that $\qit{T}(t)$ represent probabilities.
This can be achieved by using softmax activation in the final layer, or using standard normalization (i.e., dividing each output by the sum of the outputs).
It turned out that neither of the two normalizations consistently gives the best results on all datasets.
Hence we treat this as a hyper-parameter in our experiments.
 
\subsection{Training} 
We now describe the training procedure for our model defined in Eq.~\ref{eq:model}, in particular by defining an appropriate loss function.
For ease of exposition and without loss of generality, we drop the index $i$ in this section. 
Our prediction $\zthat{T}$ for a given univariate time series $(z_0, z_1, \ldots, z_{T-1})$ at time step $T$ is generated by sampling from $\qt{T}$.
So our forecast distribution for time step $T$ can be seen as sampling from the discrete random variable $\Zthat{T}$ with the probability mass function given by 
\begin{align}
f_{\Zthat{T}}(z_{t}) = \sum_{t': z_{t'} = z_{t}}\qt{T}(t'), \quad t= 0, \ldots, T-1.
\end{align}
That is, the prediction is always one of $z_t, t = 0, \ldots, T-1$ and the probability of predicting $z_t$ is the sum of the sampling probabilities of those time indices where the value $z_t$ is observed.
Similarly, the cumulative distribution function for any value $z$ is given by 
\begin{align}
	F_{\Zthat{T}}(z) = \sum_{t: z_{t} \le z}  \qt{T}(t). 
\end{align}

\textbf{Loss: Ranked Probability Score.}
Given that our forecasts are generated by sampling from the discrete random variable $\Zthat{T}$, we propose to use the ranked probability score between our probabilistic prediction (specified by $F_{\Zthat{T}}$) and the actual observation $z_{T}$.
This is essentially the sum of the quantile losses given by
\begin{align}\label{eq:rps_loss}
   \RPS(F_{\Zthat{T}}, z_{T}) = \sum_{z_t \in \{z_0, \ldots, z_{T-1}\}} \Lambda_{\alphat{t}}(z_{t}, z_{T}),
\end{align}
where $\alphat{t} = F_{\Zthat{T}}(z_{t})$ is the quantile level of $z_t$ and $\Lambda_\alpha(q, z)$ is the quantile loss given by
\[
        \Lambda_{\alpha}(q, z) = (\alpha - \indicator{z < q})(z-q).
\]

Note that the summation in the loss Eq.~\ref{eq:rps_loss} runs only on the distinct values of the past observations given by the \textit{set} $\{z_0, \ldots, z_{T-1}\}$.
The total loss of the network with parameters $\phi$ over all the training examples $\{\zit{0:T}\}$ can then be defined as
\begin{align}\label{eq:loss}
        \mathcal{L}(\phi) = \sum_{i=1}^{N} \RPS(F_{\Zithat{T}}, \zit{T})
\end{align}

\paragraph{Data Augmentation.}
Note that the loss defined in Eq. ~\ref{eq:loss} uses only a single time step $T$ to evaluate the prediction, for each training example.
Since the same model would be used for multi-step ahead prediction, we generate multiple training instances from each time series by selecting context windows with different starting time points, similar to~\citep{deepar}.
For example, assume that the training data is available from February 01 to February 21 of a daily time series and we are required to predict for 7 days.
We can define the context window to be of size, say, 14 and generate 7 training examples with the following sliding context windows: February $1 + k: 14+ k, k = 0, 1, \ldots, 6$.
For each of these seven context windows (which are passed to the network as inputs), the training loss is computed using the observations at February $15 + k, k = 0, 1, \ldots, 6$ respectively.
We do the same for all the time series in the dataset.

\subsection{Prediction}
Once the model is trained, it can generate forecast distribution for a single time step.
The multi-step ahead forecast can then be generated as described in the previous section using the same trained model.
Note that to generate multi-horizon forecasts one needs to have access to time series feature values for the future time steps, a typical assumption of global models in time series forecasting~\citep{deepar, rangapuram2018deep}.


\section{Related Work}\label{sec:rel_work}
A number of new time series forecasting methods have been proposed over the last years, in particular global deep learning methods (e.g.,~\citep{oreshkin2019n,deepar,lim2019temporal,oreshkin2020meta,rasul2021autoregressive}). 
\citet{benidis2020neural} provides a recent overview. The usage of global models~\citep{januschowski19} in forecasting has well-known predecessors (e.g.,~\citep{geweke1977dynamic} and~\citep{wang2019deepfactors} for a modern incarnation), however local models have traditionally dominated forecasting which have advantages given their parsimonious parametrization, interpretability and stemming from the fact that many time series forecasting problems consists of few time series. The surge of global models in the literature can be explained both by their theoretical superiority~\citep{montero2021principles} as well as empirical success in independent competitions~\citep{makridakis2018m4,kaggle21}. We believe that both, local and global methods will continue to have their place in forecasting and its practical application. For example, the need for fail-safe fall-back models in real-world production use-cases has been recognized~\citep{bose2017probabilistic}. Therefore, the methods presented here have both local and global versions. 

While many methods of the afore-mentioned recent global deep learning methods only consider providing point forecasts, some do provide probabilistic forecasts~\citep{Wen2017multi,deepar,rangapuram2018deep} motivated by downstream decision making problems often requiring the minimization of expected cost (e.g.,~\citep{levi2020}). The approaches to produce probabilistic forecast range from making standard parametric assumptions on the pdf (e.g.,~\citep{deepar,rabanser2020effectiveness}), to more flexible parametrizations via copulas~\citep{multivariate} or normalizing flows~\cite{rasul2020multivariate,debezenac2020}, sometimes using extensions to energy-based models~\citep{rasul2021autoregressive}, from quantile regression~\citep{Wen2017multi} to parametrization of the quantile function~\cite{aistats}. All these approaches share an inherent risk induced by potential numerical instability. For example, even estimating a standard likelihood of a linear dynamical system via a Kalman Filter (see~\citep{rangapuram2018deep} for a recent forecasting example) easily results in numerical complications unless care is taken. In contrast, the proposed methods here are almost completely fail-safe in the sense that numerical issues will not result in catastrophically wrong forecasts. While the added robustness comes at the price of some accuracy loss, the overall accuracy is nevertheless competitive and an additional benefit is the reduced amount of hyperparameter tuning necessary, even for DeepNPTS.

The general idea of constructing predictive distributions from the empirical distributions of (subsets of) observations has been explored in the context of probabilistic regression, e.g.\ in the form of Quantile Regression Forests \citep{meinshausen2006quantile} or more generally in the form of conformal prediction \citep{vovk2005}.

\section{Experiments}\label{sec:experiments}
\begin{table*}[!htbp]
	\small
	\centering
	\begin{tabular}{cccccc}
		\toprule
		{dataset} & \multicolumn{1}{p{4cm}}{\centering \numtestpoints \\ ({\predlength} $\times$ {\numrolls}) } & {\domain} & {\freq} & {\size} & {\history}\\
		\midrule
		{\exchange} & 150 (30 $\times$  5) &$\mathbb{R}^+$ & daily & 8 & 6071 \\
		{\solar} & 168 (24 $\times$ 7) &$\mathbb{R}^+$ & hourly & 137 & 7009 \\
		{\electricity} & 168 (24 $\times$ 7) &$\mathbb{R}^+$ & hourly & 370 & 5790\\
		{\traffic} & 168 (24 $\times$ 7) &$[0, 1]$ & hourly & 963 & 10413 \\
		{\taxi} & 1344 (24 $\times$ 56) &$\mathbb{N}$ & 30-min & 1214 &  1488 \\
		{\wiki} & 150 (30 $\times$ 5) &$\mathbb{N}$ & daily & 2000 & 792 \\
		\bottomrule
	\end{tabular}
	\caption{Summary of the datasets used in the evaluations: Number of time steps evaluated, data domain, frequency of observations, number of time series in the dataset, and median length of time series.}
	\label{tab:dataset_summary}
\end{table*}

We present empirical evaluation of the proposed method on the following datasets, which are publicly available via \GluonTS, a time series library~\cite{alexandrov2019gluonts}.
\begin{itemize}
	\itemsep0em
	\item \electricity: hourly time series of the electricity consumption of 370 customers \citep{Dua:2017}
	\item \exchange: daily exchange rate between 8 currencies as used in \citep{lstnet}
	\item \solar: hourly photo-voltaic production of 137 stations in Alabama State used in \citep{lstnet}
	\item \taxi: spatio-temporal traffic time series of New York taxi rides \citep{taxi:2015} taken at 1214 locations every 30 minutes in the months of January 2015 (training set) and January 2016 (test set)
	\item \traffic: hourly occupancy rate of 963 San Francisco car lanes \citep{Dua:2017}
	\item \wiki: daily page views of 9535 Wikipedia pages used in \citep{aistats}
\end{itemize}

Table~\ref{tab:dataset_summary} summarizes the key features of these datasets. 
For all datasets, the predictions are evaluated in a rolling-window fashion. 
The length of the prediction window as well as the number of such windows are given in Table~\ref{tab:dataset_summary}.
Note that for each time series in a dataset, forecasts are evaluated at a total of $\tau$ time points, where $\tau$ is equal to the length of the prediction window times the number of windows.
For a given $\tau$, the total evaluation length, let $T+\tau$ be the length of the time series available for a dataset.
Then each method initially receives time series for the first $T$ time steps which are used to tune hyperparameters in a back-test fashion, e.g., training on the first $T - \tau$ steps and validating on the last $\tau$ time steps.
Once the best hyperparameters are found, each model is once again trained on $T$ time steps and is evaluated on time steps from $T+1$ to $T+ \tau$.

\paragraph{Evaluation Criteria.} For evaluating the forecast distribution, we use the mean of quantile losses evaluated at different quantile levels ranging from $0.05$ to $0.95$ in steps of $0.05$.
Note that this is an approximate version of CRPS, a proper metric for evaluating distribution forecasts.

\paragraph{Methods Compared.} We compare against standard baselines in forecasting literature including classical models \verb|Seasonal-Naive|, \verb|ETS|, \verb|ARIMA|~\citep{hyndman2017forecasting}, \STLAR~\citep{Rforecast}, \tbats~\citep{tbats} and $\Thetaf$~\citep{Theta}, as well as deep learning based models \DeepAR~\citep{deepar} and \SQF~\citep{aistats}.
\STLAR\ first applies the STL decomposition to each time series and then fits an AR model for the seasonally adjusted data, while the seasonal component is forecast using the seasonal naive method.
\tbats\ incorporates Box-Cox transformation and Fourier representations in the state space framework to handle complex seasonal time series thereby addressing the limitations of \verb|ETS| and \verb|ARIMA|.
\Thetaf\ is an additional baseline with good empirical performance~\citep{Theta} and its forecasts are equivalent to simple exponential smoothing model with drift~\citep{unmaskingTheta}.
\DeepAR\ is an RNN-based probabilistic forecasting model that has been shown to be one of the best performing models empirically~\citep{alexandrov2019gluonts}.
\SQF\ is built on top of \DeepAR\ to output quantile function directly thereby predicting quantiles for all quantile levels~\citep{aistats}; the quantile function is modelled via a monotonic spline where the slopes and the knot positions of the spline are predicted by RNN.
Note that \SQF\ is a distribution-free method and can flexibly adapt to different output distributions.
Table~\ref{tab:hyperDeepAR} shows the tuned hyperparameter grid for \DeepAR, and \SQF.
For the \DeepNPTS\ model the hyperparameter grid is given in Table~\ref{tab:hyperDeepNPTS}.
We also include for comparison all variants of the proposed NPTS method.
In particular, we have the following four combinations: (i) \NPTS\ with and without seasonality (ii) \NPTS\ using uniform weights and exponentially decaying weights.

For the variants of \NPTS\ that use exponential kernel (\NPTS, \SeasonalNPTS) we tune the (inverse of) width parameter $\lambda$ on the validation set.
In particular, we use the grid: $\lambda \in \{1.0, 0.75, 0.5, 0.25, 0.1\}$.
For the other two variants using the uniform kernel (\Climatological, \SeasonalClimatological), there are no hyperparameters to be tuned.
The hyperparameter grid for \DeepNPTS\ is given in Table~\ref{tab:hyperDeepNPTS}.
For \verb|ETS|, \verb|ARIMA| and \tbats, we use the auto-tuning option provided by the R \verb|forecast| package~\citep{Rforecast}.
The rest do not have any hyperparameters that can be tuned.
We use the mean quantile loss as the criteria for tuning the models, since all models compared here produce distribution forecasts.

The code is publicly available in \GluonTS~\cite{alexandrov2019gluonts}.

\begin{table}[H]
	\centering
	\begin{tabular}{l|c}
		\toprule
		Parameter                     &   Range\\
		\midrule	
		\texttt{dropout}& $\{0, 0.1\}$\\
		\texttt{static feat}&   \texttt{True | Flase}        \\
		\texttt{num of RNN cells}&    ${\{40, 80\}}$     \\
		\texttt{context length} & $\{1, 2\} \times$ \texttt{pred. length}\\
		\texttt{epochs}&           $\{200, 300\}$  \\
		\midrule
		\texttt{num of spline pieces} & $\{5, 10, 15\}$ \\
		\bottomrule
	\end{tabular}
	\caption{Hyperparameter grid for \DeepAR\ (top-part) and \SQF~\cite{alexandrov2019gluonts}. \SQF\ has same parameters like \DeepAR\ except for one additional parameter shown on the last line.}
	\label{tab:hyperDeepAR}
\end{table}

\begin{table}[H]
	\centering
	\begin{tabular}{l|c}
		\toprule
		Parameter                     &   Range\\
		\midrule	
		\texttt{dropout}& $\{0, 0.1\}$\\
		\texttt{static feat}&   \texttt{True | False}        \\
		\texttt{normalization}&    \texttt{softmax, normal}     \\
		\texttt{input scaling} & \texttt{None, standardization}\\
		\texttt{loss scaling} & \texttt{None, min/max scaling}\\
		\texttt{epochs}&           $\{200, 300\}$  \\			
		\bottomrule
	\end{tabular}
	\caption{Hyperparameter grid for \DeepNPTS.}
	\label{tab:hyperDeepNPTS}
\end{table}

\paragraph{Additional Experiment Details.}
All the deep learning based methods are run using Amazon SageMaker~\citep{liberty20} on a machine with 3.4GHz processor and 32GB RAM.
The remaining methods are run on Amazon cloud instance of same configuration, 3.4GHz processor and 32GB RAM.
We used the code available in \GluonTS~\citep{alexandrov2019gluonts} forecasting library to run all the baseline methods compared.
For \DeepNPTS, the number of layers of the MLP is fixed at 2 and the number of hidden nodes (equal to the size of the context window) is chosen as a constant multiple of the prediction length.
This multiple varies for each dataset (depending on the length of the time series available) and is fixed at a large value, \textit{without tuning}, so that multiple training instances (equal to the prediction length) can be generated for each time series in the dataset.
Table~\ref{tab:context_lengths} shows the context lengths (as multiple of prediction lengths) used for different datasets.

All deep learning based methods receive time features that are automatically determined based on the frequency of the given dataset as implemented in \GluonTS.

\begin{table}[H]
	\centering{
		\begin{tabular}{l|c}
			\toprule
			dataset                     &   \texttt{context Length}\\
			\midrule	
			\wiki & $10 \times$ \texttt{pred. length}\\
			\texttt{Other datasets} & $28 \times$ \texttt{pred. length}\\
			\bottomrule
		\end{tabular}
		\caption{Context lengths used for \DeepNPTS.}
		\label{tab:context_lengths}
	}
\end{table}

\paragraph{Qualitative Analysis.}
\begin{figure*}
	\centering
	\begin{subfigure}[b]{0.5\textwidth}
		\includegraphics[width=\textwidth]{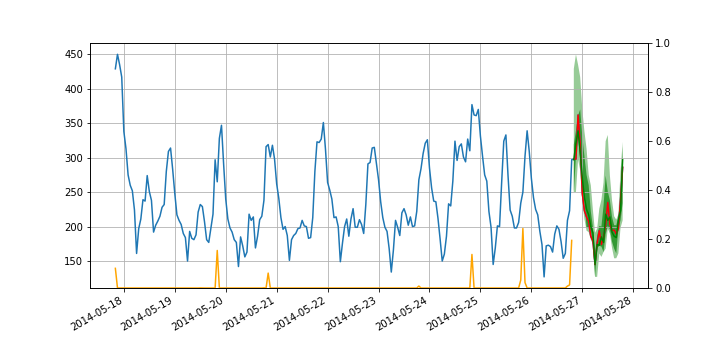}
		\caption{\electricity}
	\end{subfigure}%
	\begin{subfigure}[b]{0.5\textwidth}
		\includegraphics[width=\textwidth]{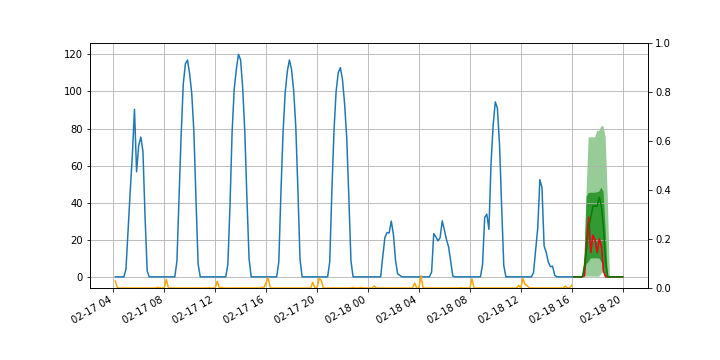}
		\caption{\solar}
	\end{subfigure}
	\begin{subfigure}[b]{0.5\textwidth}
		\includegraphics[width=\textwidth]{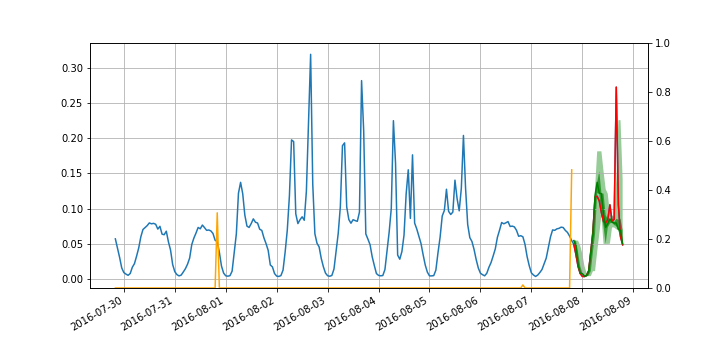}
		\caption{\traffic}
	\end{subfigure}%
	\begin{subfigure}[b]{0.5\textwidth}
		\includegraphics[width=\textwidth]{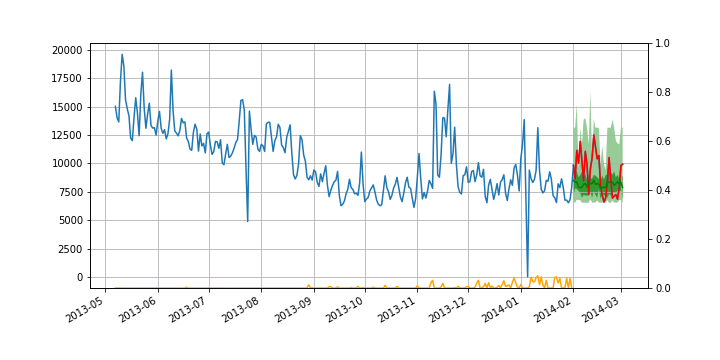}
		\caption{\wiki}
	\end{subfigure}
	\caption{Example of forecasts obtained with \DeepNPTS\ on some of the datasets used for experiments, together with the probabilities that the model outputs. The dark green line is the median forecast, surrounded by the 50\% (green) and 90\% (light green) prediction intervals. The red line is the actual target in the forecast time window. The orange line indicates the probability assigned by the model to the corresponding time point (as measured on the dual axis on the right of each plot), in making the \emph{first} time step prediction: this highlights the seasonal patterns that the model finds in the data.}
	\label{fig:predictions}
\end{figure*}

\begin{figure*}
	\centering
	\includegraphics[width=\textwidth]{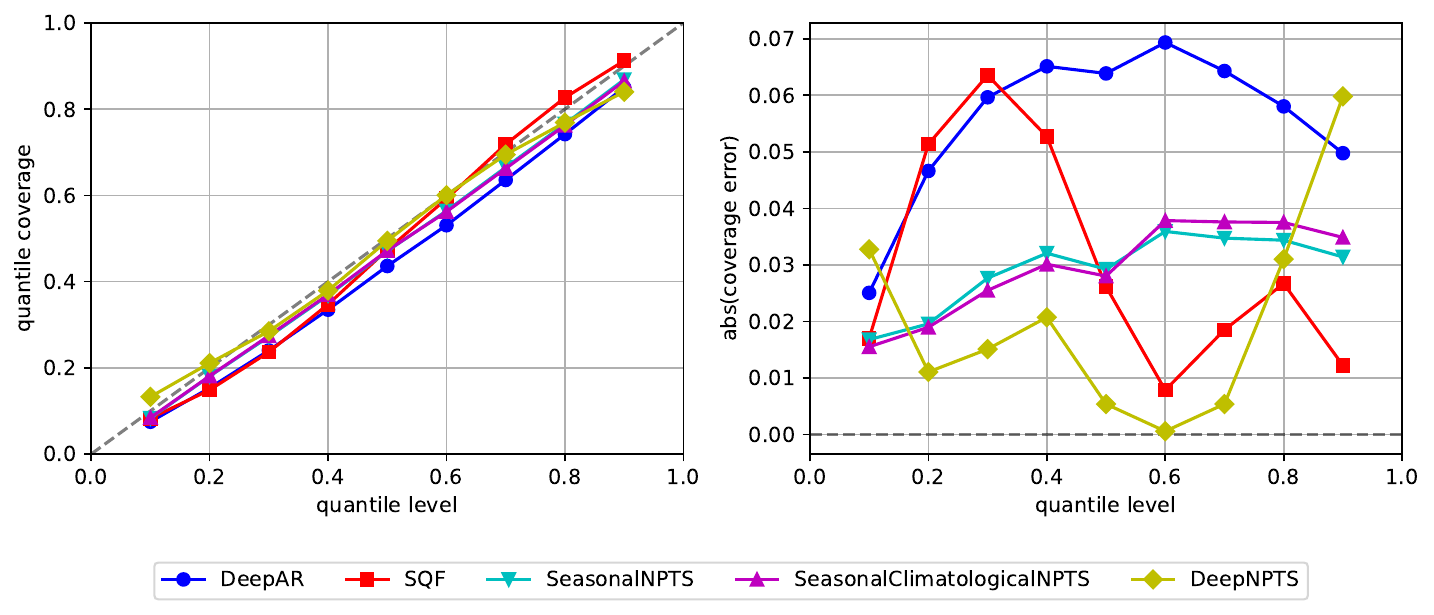}
	\caption{\small{
		Models calibration on the \traffic\ dataset. Left: calibration plot, showing what fraction of the actual data lies below the
		predicted quantiles (the ideal profile lies on the diagonal); this is a crucial quality metric for probabilistic forecasts.
		Right: calibration error (absolute value), showing the difference between coverage and quantile levels (the lower the better).
		On this example, the predictions from \DeepNPTS\ appear to be better calibrated than the ones from
		\SeasonalNPTS\ and \SeasonalClimatological, which are in turn better calibrated than the ones from \DeepAR. \SQF\ mitigated
		the calibration issue of \DeepAR\ by being distribution free method but is still worse than \DeepNPTS.
	}}
	\label{fig:calib}
\end{figure*}

Before presenting quantitative results, we first analyse the performance of \DeepNPTS\ model qualitatively by visually inspecting its forecasts and the corresponding probabilities learnt, when considering specific time series in detail (instead of overall aggregate accuracy metrics). 
Figure~\ref{fig:predictions} shows forecasts of \DeepNPTS\ for one example time series taken from each of the four datasets used in the experiments.
Each plot shows the true target as well as the 50\% and 90\% prediction intervals of the forecast distribution for the prediction window.
Additionally, we show the output of the model (i.e., sampling probabilities) with an orange line (the second axis on the right side of each plot). Note that this is the output of the model in making the prediction for the \emph{first} time step of the prediction window.
In the case of \traffic, one can notice that the last observation and the same hour on the previous week received the highest probabilities, indicating that the model correctly captures the hour-of-week seasonality pattern.
For \electricity, the same hour over multiple consecutive days was assigned high probability, indicating a hour-of-day seasonality.
For non-seasonal time series like \wiki\, the most recent time points are assigned higher probabilities.

One of the main benefits of \DeepNPTS\ is then the \textit{explainability} of its output:
although it is a deep-learning based model, one can readily explain how the model generated forecasts by looking at which time points got higher probabilities, for any given time step, and also verify if the model assigned probabilities correspond to an intuitive understanding of the data.

Next, we analyse the calibration of the forecasts of the \DeepNPTS\ model.
For a given set of \(N\) time series, denote
the predicted \(\alpha\)-quantile for the \(i\)-th series, at time \(t\), by \(\hat z_{t}^{(i)}(\alpha)\).
Then the \emph{coverage} at level \(\alpha\) is the empirical fraction of actual observations that lie below
the predicted \(\alpha\)-quantile, that is
\[
	\coverage(\alpha) = \frac{1}{NH} \sum_{
		\substack{i\in\{1,\ldots,N\},\\
		t\in\{T+1,\ldots,T+H\}}
	} \indicator{z_t^{(i)} \leq \hat z_{t}^{(i)}(\alpha)}
\]
where \(H\) denotes the prediction horizon.
This is a crucial quality metric for probabilistic forecasts:
a perfectly calibrated model has \(\coverage(\alpha) = \alpha\).
Figure~\ref{fig:calib} displays coverage for the \traffic\ dataset,
together with the calibration error that is the quantity
\(|\mbox{coverage}(\alpha) - \alpha|\).
As expected, forecasts from both the \NPTS\ and \DeepNPTS\ models are highly
calibrated in absolute terms and in relative terms, more so than \DeepAR\
with a student-t output distribution.

\begin{table*}[!htbp]


	\small
	{
		\center
		\begin{tabular}{lcccccc}

			\toprule
			dataset &      \electricity & \exchange  &     \solar &       \taxi  &          \traffic & \wiki \\
			estimator         &                  &                  &                  &                  &                  &                   \\
			\midrule
			\SeasonalNaive          &  0.070$\pm$0.000 &  0.011$\pm$0.000 &  0.605$\pm$0.000 &  0.507$\pm$0.000 &  0.251$\pm$0.000 &     0.404$\pm$0.000 \\		
			\STLAR  & \B  0.059$\pm$0.000 & \B  0.008$\pm$0.000 & 0.527$\pm$0.000 & \B  0.351$\pm$0.000 & 0.225$\pm$0.000 & 0.540$\pm$0.000\\
			\ETS  & 0.118$\pm$0.001 & \B  0.007$\pm$0.000 & \emph{1.717$\pm$0.005} & \emph{0.572$\pm$0.000} & 0.359$\pm$0.000 & \emph{0.664$\pm$0.001} \\
			\ARIMA  & \B  0.059$\pm$0.000  & \B  0.008$\pm$0.000 & 1.161$\pm$0.001 & 0.473$\pm$0.000 & 0.268$\pm$0.000 & 0.477$\pm$0.000 \\
			\Thetaf  & 0.105$\pm$0.000 & \B {0.007$\pm$0.000} & 1.083$\pm$0.000 & 0.558$\pm$0.000 & 0.331$\pm$0.000 & 0.622$\pm$0.000\\
			\tbats  & \emph{0.317$\pm$0.000} & \B  0.008$\pm$0.000 & 0.653$\pm$0.000 & {0.377$\pm$0.000} & 0.534$\pm$0.000 & 0.495$\pm$0.000\\
			\Croston & \NA & \NA & 1.402$\pm$0.000 & 0.673$\pm$0.000  & \NA & 0.334$\pm$0.000 \\  			
			\midrule
			\DeepAR                 &  \B  0.056$\pm$0.001 &  0.009$\pm$0.001 &  \B  0.426$\pm$0.004 &  \B  0.289$\pm$0.001 &  \B  0.117$\pm$0.001 &     \B  0.226$\pm$0.002 \\		
			\SQF    &  \B  0.056$\pm$0.003 &  0.009$\pm$0.001 &  \B  0.390$\pm$0.004 &  \B  0.283$\pm$0.001 &  \B  0.113$\pm$0.002 &   \B  {0.235$\pm$0.012} \\			
			\midrule
			\Climatological         &  0.230$\pm$0.001 &  \emph{0.026$\pm$0.000} &  0.805$\pm$0.001 &  0.473$\pm$0.000 &  0.403$\pm$0.000 &     0.291$\pm$0.000 \\		
			\SeasonalClimatological &  0.061$\pm$0.000 &  \emph{0.026$\pm$0.000} &  0.791$\pm$0.000 &  0.474$\pm$0.000 &  0.174$\pm$0.000 &     0.285$\pm$0.000 \\
			\NPTS                   &  0.230$\pm$0.001 &  0.021$\pm$0.000 &  0.802$\pm$0.001 &  0.473$\pm$0.000 &  0.403$\pm$0.000 &     0.277$\pm$0.000 \\
			\SeasonalNPTS           &  0.060$\pm$0.000 &  0.020$\pm$0.000 &  0.790$\pm$0.000 &  0.474$\pm$0.000 &  0.173$\pm$0.000 &     0.269$\pm$0.000 \\		
			\DeepNPTS               &  \B  0.059$\pm$0.004 &  0.009$\pm$0.000 &  \B  0.447$\pm$0.011 &  {0.413$\pm$0.015} &  \B  0.163$\pm$0.036 &     \B  0.232$\pm$0.000 \\	
			\bottomrule
		\end{tabular}
		\caption{Mean quantile losses averaged over 5 runs (lower is better). Best three methods are highlighted in \textbf{bold}, the worst method is in \textit{italic}. The top part shows the results of local models, while the middle part show the same for global deep learning models. The bottom half shows different variants of the proposed \NPTS\ model. \label{tab:CRPS}
		}
	}
\end{table*}

\paragraph{Quantitative Results.}
Table~\ref{tab:CRPS} summarizes the quantitative results of the datasets considered via mean quantile losses, averaged over 5 runs. 
The top three best performing methods are highlighted with \textbf{boldface}.
The top section shows the baselines considered and the bottom section shows different variants of the proposed NPTS method.
First note that \DeepNPTS\ model always achieves much better results than any other variant of \NPTS\ showing that learning the sampling strategy clearly helps.
Moreover, \DeepNPTS\ comes as one of the top three methods in 4 out of 6 datasets considered and in the remaining its results are close to the best performing methods.
Importantly for the \textit{difficult} datasets like \solar\ (non-negative data with several zeros), \traffic\ (data lies in $(0, 1)$) and \wiki\ (integer data), the gap between the performance of \DeepNPTS\ and the standard baselines like \ETS\ and \ARIMA\ is very high; see the next paragraph for the visualization of distribution of these datasets.
Even some of the other variants of \NPTS\ (with fixed sampling strategy) performed better than \ETS\ and \ARIMA\ in these datasets without any training (other than the tuning of $\lambda$). \DeepNPTS\ is particularly robust since it is never among the worst methods. 
All the variants of \NPTS\ run much faster than the standard baselines \ETS,\ \ARIMA\ and \tbats, which fit a different model to each time series in the dataset.
These observations, in addition to explainability and being able to generate calibrated forecasts, further support our claim that the proposed methods in general and \DeepNPTS\ in particular, are good baselines to consider for arbitrary data distributions and can be safely used as fast, fall-back methods in practical applications. 

\begin{figure*}
	\centering
	\includegraphics[width=\textwidth]{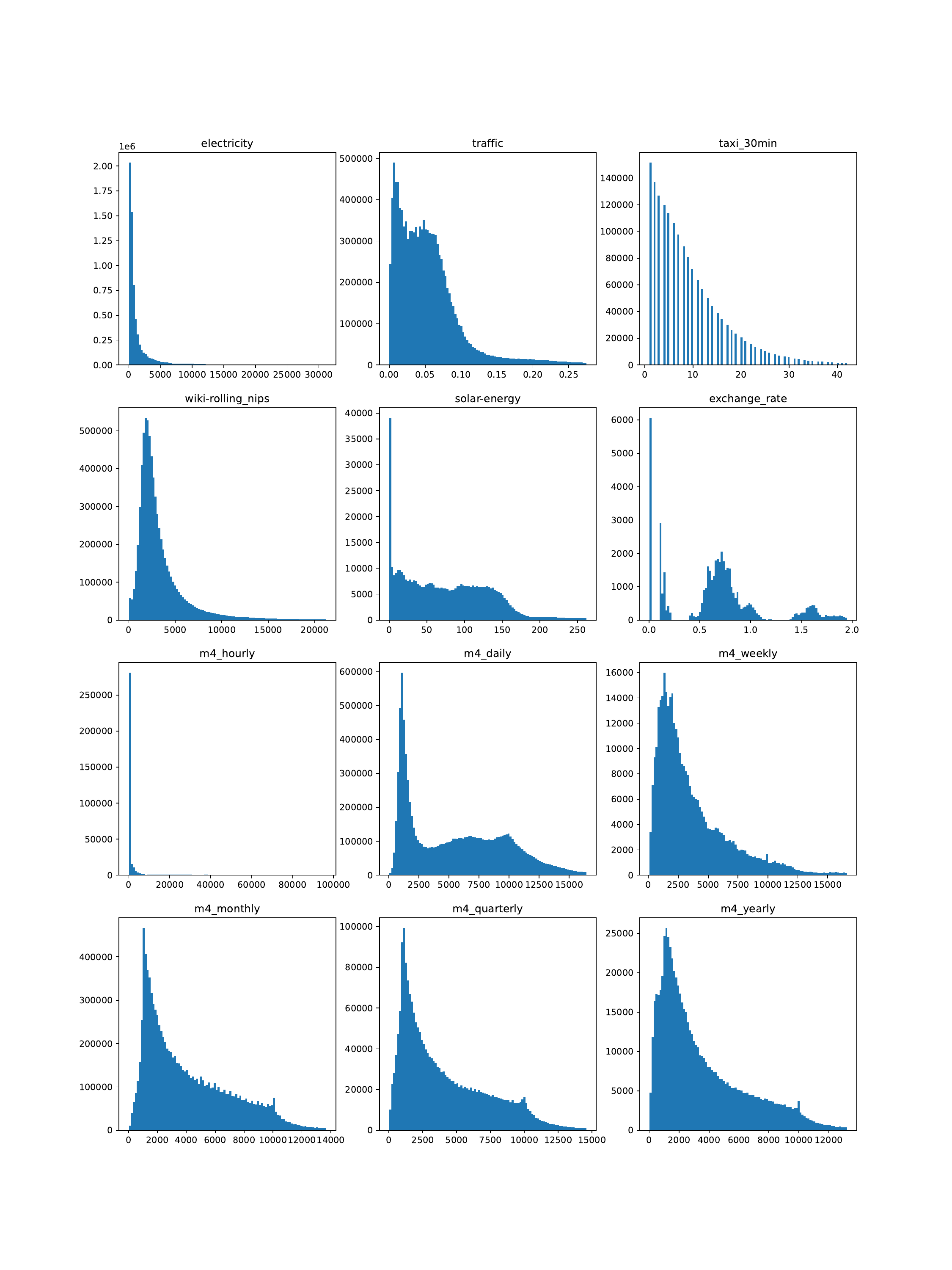}
	\caption{Histogram of observed values of the (training part of) time series for all datasets used in the evaluations.}
	\label{fig:distr}
\end{figure*}

\paragraph{Data Visualization.}
To illustrate the diversity of the datasets used, we plot the distribution of observed values in the training part of the time series marginalized over time and item dimensions (each dataset contains time series corresponding to different \textit{items}).
The plot is shown in Figure~\ref{fig:distr} for all the datasets.

\section{Conclusion}\label{sec:conclusion}
In this paper we presented a novel probabilistic forecasting method, in both an extremely simple yet effective local version, and an adaptive, deep-learning-based global version. In both variants, the proposed methods can serve as robust, fail-safe forecasting methods that are able to provide accurate probabilistic forecasts. We achieve robustness by constructing the predictive distributions by reweighting the empirical distribution of the past observations, and achieve accuracy by taking advantage of learning the context-dependent weighting globally, across time series. We show in empirical evaluations that the methods, NPTS and DeepNPTS, perform roughly on-par with the  state-of-the-art deep-learning based methods both quantitatively and qualitatively.

\bibliographystyle{abbrvnat}
\bibliography{references}

\begin{thebibliography}{45}
\providecommand{\natexlab}[1]{#1}
\providecommand{\url}[1]{\texttt{#1}}
\expandafter\ifx\csname urlstyle\endcsname\relax
  \providecommand{\doi}[1]{doi: #1}\else
  \providecommand{\doi}{doi: \begingroup \urlstyle{rm}\Url}\fi

\bibitem[Akram et~al.(2009)Akram, Hyndman, and Ord]{Akram2008}
M.~Akram, J.~Hyndman, and K.~Ord.
\newblock Exponential smoothing and non-negative data.
\newblock 51\penalty0 (4):\penalty0 415--432, 2009.

\bibitem[Alexandrov et~al.(2020)Alexandrov, Benidis, Bohlke-Schneider,
  Flunkert, Gasthaus, Januschowski, Maddix, Rangapuram, Salinas, and
  Schulz]{alexandrov2019gluonts}
Alexander Alexandrov, Konstantinos Benidis, Michael Bohlke-Schneider, Valentin
  Flunkert, Jan Gasthaus, Tim Januschowski, Danielle~C Maddix, Syama
  Rangapuram, David Salinas, and Jasper Schulz.
\newblock {GluonTS: Probabilistic Time Series Models in Python}.
\newblock \emph{Journal of Machine Learning Research}, 21\penalty0
  (116):\penalty0 1--6, 2020.

\bibitem[Assimakopoulos and Nikolopoulos(2000)]{Theta}
V.~Assimakopoulos and K.~Nikolopoulos.
\newblock {The theta model: a decomposition approach to forecasting}.
\newblock \emph{International Journal of Forecasting}, 16\penalty0
  (4):\penalty0 521--530, 2000.

\bibitem[Benidis et~al.(2020)Benidis, Rangapuram, Flunkert, Wang, Maddix,
  Turkmen, Gasthaus, Bohlke-Schneider, Salinas, Stella, Callot, and
  Januschowski]{benidis2020neural}
Konstantinos Benidis, Syama~Sundar Rangapuram, Valentin Flunkert, Bernie Wang,
  Danielle Maddix, Caner Turkmen, Jan Gasthaus, Michael Bohlke-Schneider, David
  Salinas, Lorenzo Stella, Laurent Callot, and Tim Januschowski.
\newblock Neural forecasting: Introduction and literature overview, 2020.

\bibitem[Bojer and Meldgaard(2021)]{kaggle21}
Casper~Solheim Bojer and Jens~Peder Meldgaard.
\newblock Kaggle forecasting competitions: An overlooked learning opportunity.
\newblock \emph{International Journal of Forecasting}, 37\penalty0
  (2):\penalty0 587--603, 2021.
\newblock ISSN 0169-2070.

\bibitem[B{\"o}se et~al.(2017)B{\"o}se, Flunkert, Gasthaus, Januschowski,
  Lange, Salinas, Schelter, Seeger, and Wang]{bose2017probabilistic}
Joos-Hendrik B{\"o}se, Valentin Flunkert, Jan Gasthaus, Tim Januschowski,
  Dustin Lange, David Salinas, Sebastian Schelter, Matthias Seeger, and Yuyang
  Wang.
\newblock Probabilistic demand forecasting at scale.
\newblock \emph{Proceedings of the VLDB Endowment}, 10\penalty0 (12):\penalty0
  1694--1705, 2017.

\bibitem[Chen and Guestrin(2016)]{Chen2016}
Tianqi Chen and Carlos Guestrin.
\newblock {XGBoost}: A scalable tree boosting system.
\newblock In \emph{Proceedings of the 22nd ACM SIGKDD International Conference
  on Knowledge Discovery and Data Mining}, KDD '16, pages 785--794. ACM, 2016.

\bibitem[Croston(1972)]{Croston1972}
J.~Croston.
\newblock Forecasting and stock control for intermittent demands.
\newblock \emph{Journal of Operational Research Quarterly}, 23\penalty0
  (3):\penalty0 289--303, 1972.

\bibitem[de~B\'{e}zenac et~al.(2020)de~B\'{e}zenac, Rangapuram, Benidis,
  Bohlke-Schneider, Kurle, Stella, Hasson, Gallinari, and
  Januschowski]{debezenac2020}
Emmanuel de~B\'{e}zenac, Syama~Sundar Rangapuram, Konstantinos Benidis, Michael
  Bohlke-Schneider, Richard Kurle, Lorenzo Stella, Hilaf Hasson, Patrick
  Gallinari, and Tim Januschowski.
\newblock Normalizing kalman filters for multivariate time series analysis.
\newblock In \emph{Advances in Neural Information Processing Systems},
  volume~33, pages 2995--3007, 2020.

\bibitem[Dheeru and Karra~Taniskidou(2017)]{Dua:2017}
Dua Dheeru and Efi Karra~Taniskidou.
\newblock {UCI} machine learning repository.
\newblock \url{http://archive.ics.uci.edu/ml}, 2017.

\bibitem[Epstein(1969)]{Epstein:1969}
E.~S. Epstein.
\newblock A scoring system for probability forecasts of ranked categories.
\newblock \emph{J. Appl. Meteor.}, 8:\penalty0 985--987, 1969.

\bibitem[Gasthaus et~al.(2019)Gasthaus, Benidis, Wang, Rangapuram, Salinas,
  Flunkert, and Januschowski]{aistats}
Jan Gasthaus, Konstantinos Benidis, Yuyang Wang, Syama~S. Rangapuram, David
  Salinas, Valentin Flunkert, and Tim Januschowski.
\newblock Probabilistic forecasting with spline quantile function {RNNs}.
\newblock \emph{AISTATS}, 2019.

\bibitem[Geweke(1977)]{geweke1977dynamic}
John Geweke.
\newblock The dynamic factor analysis of economic time series.
\newblock \emph{Latent variables in socio-economic models}, 1977.

\bibitem[Gneiting and Raftery(2007)]{gneiting2007strictly}
Tilmann Gneiting and Adrian~E Raftery.
\newblock Strictly proper scoring rules, prediction, and estimation.
\newblock \emph{Journal of the American Statistical Association}, 102\penalty0
  (477):\penalty0 359--378, 2007.

\bibitem[Hyndman et~al.(2008)Hyndman, Koehler, Ord, and Snyder]{Hyndman2008}
Rob Hyndman, Anne Koehler, Keith Ord, and Ralph Snyder.
\newblock \emph{Forecasting with exponential smoothing. The state space
  approach}.
\newblock 2008.
\newblock \doi{10.1007/978-3-540-71918-2}.

\bibitem[Hyndman and Athanasopoulos(2017)]{hyndman2017forecasting}
Rob~J Hyndman and George Athanasopoulos.
\newblock Forecasting: Principles and practice.
\newblock \emph{\url{www.otexts.org/fpp}}, 987507109, 2017.

\bibitem[Hyndman and Billah(2003)]{unmaskingTheta}
Rob~J. Hyndman and Baki Billah.
\newblock {Unmasking the Theta method}.
\newblock \emph{International Journal of Forecasting}, 19\penalty0
  (2):\penalty0 287--290, 2003.

\bibitem[Hyndman and Khandakar(2008)]{Rforecast}
Rob~J Hyndman and Yeasmin Khandakar.
\newblock Automatic time series forecasting: the forecast package for {R}.
\newblock \emph{Journal of Statistical Software}, 26\penalty0 (3):\penalty0
  1--22, 2008.
\newblock URL \url{https://www.jstatsoft.org/article/view/v027i03}.

\bibitem[Januschowski et~al.(2019)Januschowski, Gasthaus, Wang, Salinas,
  Flunkert, Bohlke-Schneider, and Callot]{januschowski19}
Tim Januschowski, Jan Gasthaus, Yuyang Wang, David Salinas, Valentin Flunkert,
  Michael Bohlke-Schneider, and Laurent Callot.
\newblock Criteria for classifying forecasting methods.
\newblock \emph{International Journal of Forecasting}, 2019.

\bibitem[Ke et~al.(2017)Ke, Meng, Finley, Wang, Chen, Ma, Ye, and
  Liu]{LightGBM17}
Guolin Ke, Qi~Meng, Thomas Finley, Taifeng Wang, Wei Chen, Weidong Ma, Qiwei
  Ye, and Tie-Yan Liu.
\newblock Light{GBM}: A highly efficient gradient boosting decision tree.
\newblock In I.~Guyon, U.~V. Luxburg, S.~Bengio, H.~Wallach, R.~Fergus,
  S.~Vishwanathan, and R.~Garnett, editors, \emph{Advances in Neural
  Information Processing Systems}, volume~30. Curran Associates, Inc., 2017.

\bibitem[Lai et~al.(2017)Lai, Chang, Yang, and Liu]{lstnet}
Guokun Lai, Wei{-}Cheng Chang, Yiming Yang, and Hanxiao Liu.
\newblock Modeling long- and short-term temporal patterns with deep neural
  networks.
\newblock \emph{CoRR}, abs/1703.07015, 2017.

\bibitem[Liberty et~al.(2020)Liberty, Karnin, Xiang, Rouesnel, Coskun,
  Nallapati, Delgado, Sadoughi, Astashonok, Das, Balioglu, Chakravarty, Jha,
  Gautier, Arpin, Januschowski, Flunkert, Wang, Gasthaus, Stella, Rangapuram,
  Salinas, Schelter, and Smola]{liberty20}
Edo Liberty, Zohar Karnin, Bing Xiang, Laurence Rouesnel, Baris Coskun, Ramesh
  Nallapati, Julio Delgado, Amir Sadoughi, Yury Astashonok, Piali Das, Can
  Balioglu, Saswata Chakravarty, Madhav Jha, Philip Gautier, David Arpin, Tim
  Januschowski, Valentin Flunkert, Yuyang Wang, Jan Gasthaus, Lorenzo Stella,
  Syama Rangapuram, David Salinas, Sebastian Schelter, and Alex Smola.
\newblock Elastic machine learning algorithms in amazon sagemaker.
\newblock In \emph{Proceedings of the 2020 ACM SIGMOD International Conference
  on Management of Data}, SIGMOD '20, page 731–737, New York, NY, USA, 2020.
  Association for Computing Machinery.
\newblock ISBN 9781450367356.

\bibitem[Lim et~al.(2019)Lim, Arik, Loeff, and Pfister]{lim2019temporal}
Bryan Lim, Sercan~O. Arik, Nicolas Loeff, and Tomas Pfister.
\newblock Temporal fusion transformers for interpretable multi-horizon time
  series forecasting, 2019.

\bibitem[Livera et~al.(2011)Livera, Hyndman, and Snyder]{tbats}
Alysha M.~De Livera, Rob~J. Hyndman, and Ralph~D. Snyder.
\newblock Forecasting time series with complex seasonal patterns using
  exponential smoothing.
\newblock \emph{Journal of the American Statistical Association}, 106\penalty0
  (496):\penalty0 1513--1527, 2011.

\bibitem[Makridakis et~al.(2018)Makridakis, Spiliotis, and
  Assimakopoulos]{makridakis2018m4}
Spyros Makridakis, Evangelos Spiliotis, and Vassilios Assimakopoulos.
\newblock The {M4} competition: Results, findings, conclusion and way forward.
\newblock \emph{International Journal of Forecasting}, 34\penalty0
  (4):\penalty0 802--808, 2018.

\bibitem[Matheson and Winkler(1976)]{matheson1976scoring}
James~E Matheson and Robert~L Winkler.
\newblock Scoring rules for continuous probability distributions.
\newblock \emph{Management science}, 22\penalty0 (10):\penalty0 1087--1096,
  1976.

\bibitem[Meinshausen(2006)]{meinshausen2006quantile}
Nicolai Meinshausen.
\newblock Quantile regression forests.
\newblock \emph{Journal of Machine Learning Research}, 7\penalty0
  (Jun):\penalty0 983--999, 2006.

\bibitem[Montero-Manso and Hyndman(2021)]{montero2021principles}
Pablo Montero-Manso and Rob~J Hyndman.
\newblock Principles and algorithms for forecasting groups of time series:
  Locality and globality.
\newblock \emph{International Journal of Forecasting}, 2021.

\bibitem[Oreshkin et~al.(2019)Oreshkin, Carpov, Chapados, and
  Bengio]{oreshkin2019n}
Boris~N Oreshkin, Dmitri Carpov, Nicolas Chapados, and Yoshua Bengio.
\newblock N-beats: Neural basis expansion analysis for interpretable time
  series forecasting.
\newblock \emph{arXiv preprint arXiv:1905.10437}, 2019.

\bibitem[Oreshkin et~al.(2020)Oreshkin, Carpov, Chapados, and
  Bengio]{oreshkin2020meta}
Boris~N Oreshkin, Dmitri Carpov, Nicolas Chapados, and Yoshua Bengio.
\newblock Meta-learning framework with applications to zero-shot time-series
  forecasting.
\newblock \emph{arXiv preprint arXiv:2002.02887}, 2020.

\bibitem[Petropoulos et~al.(2021)Petropoulos, Apiletti, Assimakopoulos,
  Zied~Babai, Barrow, and Ben~Taieb]{petropoulos2021forecasting}
Fotios Petropoulos, Daniele Apiletti, Vassilios Assimakopoulos, Mohamed
  Zied~Babai, Devon~K. Barrow, and Souhaib Ben~Taieb.
\newblock Forecasting: theory and practice.
\newblock \emph{International Journal of Forecasting}, 2021.

\bibitem[Rabanser et~al.(2020)Rabanser, Januschowski, Flunkert, Salinas, and
  Gasthaus]{rabanser2020effectiveness}
Stephan Rabanser, Tim Januschowski, Valentin Flunkert, David Salinas, and Jan
  Gasthaus.
\newblock The effectiveness of discretization in forecasting: An empirical
  study on neural time series models.
\newblock \emph{arXiv preprint arXiv:2005.10111}, 2020.

\bibitem[Rangapuram et~al.(2018)Rangapuram, Seeger, Gasthaus, Stella, Wang, and
  Januschowski]{rangapuram2018deep}
Syama~Sundar Rangapuram, Matthias~W Seeger, Jan Gasthaus, Lorenzo Stella,
  Yuyang Wang, and Tim Januschowski.
\newblock Deep state space models for time series forecasting.
\newblock In \emph{{Advances in Neural Information Processing Systems}}, pages
  7785--7794, 2018.

\bibitem[Raschka(2021)]{Raschka}
S.~Raschka.
\newblock Machine learning {FAQ}. {W}hat is the difference between a parametric
  learning algorithm and a nonparametric learning algorithm?
\newblock
  \url{https://sebastianraschka.com/faq/docs/parametric_vs_nonparametric.html},
  2021.

\bibitem[Rasul et~al.(2020)Rasul, Sheikh, Schuster, Bergmann, and
  Vollgraf]{rasul2020multivariate}
Kashif Rasul, Abdul-Saboor Sheikh, Ingmar Schuster, Urs Bergmann, and Roland
  Vollgraf.
\newblock Multi-variate probabilistic time series forecasting via conditioned
  normalizing flows.
\newblock \emph{arXiv preprint arXiv:2002.06103}, 2020.

\bibitem[Rasul et~al.(2021)Rasul, Seward, Schuster, and
  Vollgraf]{rasul2021autoregressive}
Kashif Rasul, Calvin Seward, Ingmar Schuster, and Roland Vollgraf.
\newblock Autoregressive denoising diffusion models for multivariate
  probabilistic time series forecasting.
\newblock \emph{arXiv preprint arXiv:2101.12072}, 2021.

\bibitem[Salinas et~al.(2019{\natexlab{a}})Salinas, Bohlke-Schneider, Callot,
  Medico, and Gasthaus]{multivariate}
David Salinas, Michael Bohlke-Schneider, Laurent Callot, Roberto Medico, and
  Jan Gasthaus.
\newblock High-dimensional multivariate forecasting with low-rank gaussian
  copula processes.
\newblock In \emph{{Advances in Neural Information Processing Systems 32}},
  2019{\natexlab{a}}.

\bibitem[Salinas et~al.(2019{\natexlab{b}})Salinas, Flunkert, Gasthaus, and
  Januschowski]{deepar}
David Salinas, Valentin Flunkert, Jan Gasthaus, and Tim Januschowski.
\newblock Deep{AR}: Probabilistic forecasting with autoregressive recurrent
  networks.
\newblock \emph{International Journal of Forecasting}, 2019{\natexlab{b}}.

\bibitem[Seeger et~al.(2016)Seeger, Salinas, and Flunkert]{seeger16}
Matthias~W Seeger, David Salinas, and Valentin Flunkert.
\newblock Bayesian intermittent demand forecasting for large inventories.
\newblock In D.~Lee, M.~Sugiyama, U.~Luxburg, I.~Guyon, and R.~Garnett,
  editors, \emph{Advances in Neural Information Processing Systems}, volume~29.
  Curran Associates, Inc., 2016.
\newblock URL
  \url{https://proceedings.neurips.cc/paper/2016/file/03255088ed63354a54e0e5ed957e9008-Paper.pdf}.

\bibitem[Simchi-Levi and Simchi-Levi(2020)]{levi2020}
David Simchi-Levi and Edith Simchi-Levi.
\newblock We need a stress test for critical supply chains.
\newblock \emph{Harvard Business Review}, 2020.

\bibitem[Snyder et~al.(2012)Snyder, Ord, and Beaumont]{Snyder12}
Ralph~D. Snyder, J.~Keith Ord, and Adrian Beaumont.
\newblock {Forecasting the intermittent demand for slow-moving inventories: A
  modelling approach}.
\newblock \emph{International Journal of Forecasting}, 28\penalty0
  (2):\penalty0 485--496, 2012.

\bibitem[Taxi and Commission(2015)]{taxi:2015}
NYC Taxi and Limousine Commission.
\newblock {TLC} trip record data.
\newblock \url{https://www1.nyc.gov/site/tlc/about/tlc-trip-record-data.page},
  2015.

\bibitem[Vovk et~al.(2005)Vovk, Gammerman, and Shafer]{vovk2005}
Vladimir Vovk, Alex Gammerman, and Glenn Shafer.
\newblock \emph{Algorithmic Learning in a Random World}.
\newblock Springer-Verlag, Berlin, Heidelberg, 2005.
\newblock ISBN 0387001522.

\bibitem[Wang et~al.(2019)Wang, Smola, Maddix, Gasthaus, Foster, and
  Januschowski]{wang2019deepfactors}
Yuyang Wang, Alex Smola, Danielle Maddix, Jan Gasthaus, Dean Foster, and Tim
  Januschowski.
\newblock Deep factors for forecasting.
\newblock In \emph{{International Conference on Machine Learning}}, pages
  6607--6617, 2019.

\bibitem[Wen et~al.(2017)Wen, Torkkola, and Narayanaswamy]{Wen2017multi}
Ruofeng Wen, Kari Torkkola, and Balakrishnan Narayanaswamy.
\newblock A multi-horizon quantile recurrent forecaster.
\newblock \emph{arXiv preprint arXiv:1711.11053}, 2017.

\end{thebibliography}

\end{document}